\documentclass[10pt,twocolumn,letterpaper]{article}

\usepackage{cvpr}
\usepackage{times}
\usepackage{epsfig}
\usepackage{graphicx}
\usepackage{amsmath}
\usepackage{amssymb}
\usepackage{subfig}
\usepackage{authblk}
\usepackage{cite}
\usepackage{comment}
\usepackage{adjustbox}
\usepackage{textcomp}
\usepackage{array,multirow,graphicx}
\usepackage[table]{xcolor}
\usepackage{colortbl}
\usepackage{bbm}
\usepackage{makecell}
\usepackage[percent]{overpic}

\usepackage{booktabs}
\usepackage{arydshln}

\usepackage[breaklinks=true,bookmarks=false]{hyperref}

\cvprfinalcopy % *** Uncomment this line for the final submission

\def\cvprPaperID{****} % *** Enter the CVPR Paper ID here

\begin{document}

%%%%%%%%% TITLE
\title{Dual Attention Network for Scene Segmentation}
\author[$^{1,3}$ ]{
Jun Fu
}
\author[$^{1}$]{
\; Jing Liu\thanks{Corresponding Author}
}
\author[$^{1}$]{
\; Haijie Tian 
}
\author[$^{2}$]{
\; Yong Li 
}

\author[$^{2}$]{
\\Yongjun Bao
}
\author[$^{1,3}$]{
\;Zhiwei Fang
}
\author[$^{1}$]{
\;Hanqing Lu
}

\affil[ ]{$^1$National Laboratory of Pattern Recognition, Institute of Automation, Chinese Academy of Sciences\; $^2 $Business Growth BU, JD.com\; $^3$ University of Chinese Academy of Sciences}
\affil[ ]{ 
\tt\small \{jun.fu,jliu,zhiwei.fang,luhq\}@nlpr.ia.ac.cn,hjtian\_bit@163.com,\{liyong5,baoyongjun\}@jd.com}

\renewcommand\Authsep{  } 
\renewcommand\Authands{  }

\maketitle
\thispagestyle{empty}

%%%%%%%%% ABSTRACT
\begin{abstract}
In this paper, we address  the scene segmentation task by 
capturing rich contextual  dependencies based on the self-attention  mechanism. Unlike previous works that capture contexts  by multi-scale feature fusion, we propose a Dual Attention Network (DANet) to adaptively integrate local features with their global dependencies.
Specifically, we append two types of attention modules  on top of dilated FCN, which model the semantic interdependencies in spatial and channel dimensions respectively.
The position attention module selectively aggregates the
feature at each position  by a weighted sum of the features at all positions. Similar features would be related to each other regardless of their distances.  Meanwhile, the channel attention module selectively emphasizes interdependent channel maps by integrating associated features among all  channel maps.
We  sum the outputs of  the two attention modules to further improve feature representation which contributes to  more precise segmentation results. We achieve new state-of-the-art segmentation performance on three challenging scene segmentation datasets, i.e., Cityscapes, PASCAL Context and COCO Stuff dataset. In particular, a Mean IoU score of 81.5\% on Cityscapes test set is achieved without using  coarse data.\footnote{Links can be found at \url{https://github.com/junfu1115/DANet/}}. 
\end{abstract}

%%%%%%%%% BODY TEXT
\section{Introduction}

Scene segmentation is  a fundamental and challenging problem, whose goal is to segment and parse a scene image into different image regions associated with semantic categories including stuff (e.g. sky, road, grass)  and discrete objects  (e.g. person, car, bicycle). 
The study of this task can be applied to potential applications, such as automatic driving, robot sensing and image editing. In order to accomplish the task of scene segmentation effectively, we need to distinguish some confusing categories and take into account objects with different appearance. For example, regions of '‘field’' and '‘grass’' are often indistinguishable, and the objects of '‘cars'‘ may often be affected by scales, occlusion and illumination. Therefore, it is necessary to enhance the discriminative ability of feature  representations for pixel-level recognition. 

\begin{figure}[!t]
        \centering
        \includegraphics[width=1\linewidth]{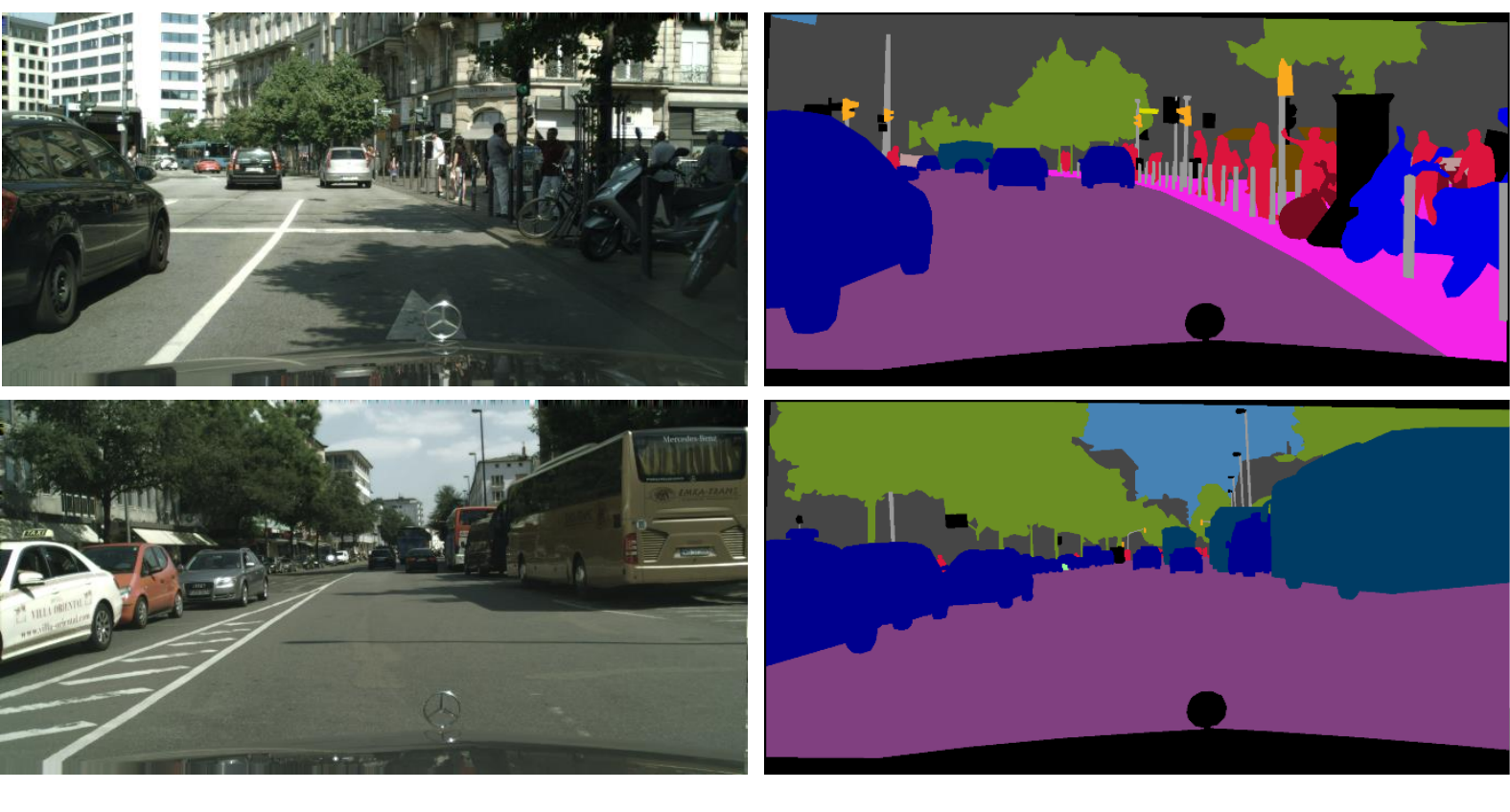}
        \vspace{-1em}
        \caption{The goal of scene segmentation is to recognize each pixel including stuff, diverse objects. The various scales, occlusion and illumination changing of objects/stuff make it challenging to parsing each pixel.}
         \label{example}%
        \vspace{-1em}
\end{figure}

Recently, state-of-the-art methods based on Fully Convolutional Networks (FCNs) \cite{FCN} have been  proposed to address the above issues.
One way is to  utilize the multi-scale context fusion. For example, some works \cite{deeplabv3,PSPNet,deeplabv2} aggregate multi-scale contexts via combining feature maps generated by  different dilated convolutions and pooling operations. And some works \cite{Peng2017LargeKM,encnet} capture richer global context information by enlarging the kernel size with a decomposed structure or introducing an effective  encoding layer on top of the network. In addition,  the encoder-decoder structures \cite{refinenet,U-net,ding2018context} are proposed to fuse mid-level and high-level semantic features.
Although the  context fusion helps to  capture different scales objects, it can not  leverage  the relationship between objects or stuff in a global view,  which is also essential to scene segmentation.

Another type of methods employs recurrent neural networks to  exploit long-range dependencies, thus improving scene segmentation accuracy. The method based on 2D LSTM networks \cite{SceneLabeling} is proposed to capture complex spatial dependencies on labels. The work \cite{DAGRNN} builds a  recurrent neural network  with directed acyclic graph to capture the rich contextual dependencies over local features.  However, these methods capture the global relationship implicitly with recurrent neural networks, whose effectiveness relies heavily on the learning outcome of the long-term memorization.

\begin{figure*}[!t]
        \centering
        \includegraphics[width=1\linewidth]{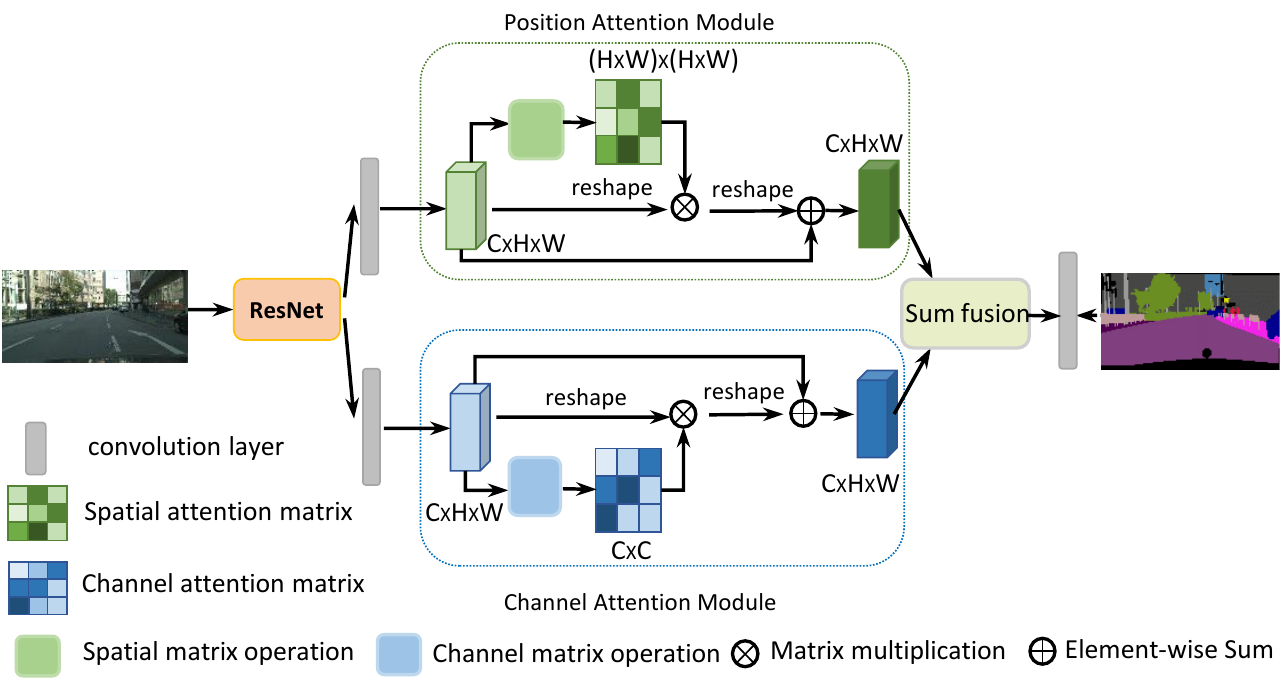}
        \vspace{-1em}
        \caption{An overview of the Dual Attention Network. (Best viewed in color)}
         \label{DAN}%
        \vspace{-1em}
\end{figure*}

To address above problems,  we propose a novel framework, called as Dual  Attention Network (DANet), for natural scene image segmentation, which is illustrated in Figure. \ref{DAN}. It introduces a self-attention mechanism  to capture   features dependencies in the spatial and channel dimensions respectively. Specifically, we append two parallel attention modules on top of dilated FCN. One is a \emph{position attention module} and the other is a \emph{channel attention module}.  
For the position attention module, we introduce  the self-attention mechanism to capture the spatial dependencies  between any two positions of the feature maps.
For the feature at a certain position, it is updated via aggregating features at all positions with weighted summation, where the weights are decided by the feature similarities between the corresponding two positions. That is, any two positions with similar features can contribute mutual improvement regardless of their distance in spatial dimension.
For the channel attention module,  we use the similar self-attention mechanism to capture the channel dependencies  between any two channel maps, and update each channel map with a weighted sum of all channel maps.  Finally, the outputs of these two attention modules are fused to  further enhance the feature representations.  

It should be noted that our method  is more effective and flexible than previous  methods \cite{deeplabv3,PSPNet} when dealing with  complex and diverse scenes. Take the street scene in Figure. \ref{example}  as an example. 
First, some '‘person'‘ and  '‘traffic light'‘ in the first row are  inconspicuous or incomplete objects due to lighting and view. If simple contextual embedding is explored, the context from dominated salient objects (e.g. car, building) would harm those inconspicuous object labeling. By contrast, our attention model selectively aggregates  the similar features of  inconspicuous objects  to highlight  their feature representations  and avoid the influence of  salient objects.
Second, the scales of the '‘car'‘ and   '‘person'‘ are diverse,   and recognizing such diverse objects requires contextual information at different scales. That is, the features at different scale should be treated equally to represent the same semantics. Our model with attention mechanism just aims to adaptively integrate similar features at any scales from a global view, and this can solve the above problem to some extent.
Third, we explicitly take  spatial and channel relationships into consideration, so that scene understanding could benefit from long-range dependencies.

Our main contributions can be summarized as follows:

\begin{itemize}
\item We propose a novel  Dual Attention Network (DANet) with  self-attention mechanism to  enhance the discriminant ability of feature  representations for scene segmentation.

\item A position attention module is proposed to learn the spatial interdependencies of features and a  channel attention module  is designed to model  channel interdependencies. It significantly improves the segmentation results by  modeling rich contextual dependencies  over local features.

\item We  achieve new state-of-the-art results on three popular benchmarks including Cityscapes dataset \cite{cityscapes},  PASCAL Context dataset \cite{pcontext} and COCO Stuff  dataset \cite{cocostuff}. 
\end{itemize}

%-------------------------------------------------------------------------

\section{Related Work}

\noindent{\bfseries Semantic Segmentation.} Fully Convolutional Networks (FCNs) based methods have made great progress in semantic  segmentation. There are several model variants proposed to enhance contextual aggregation.  
First,  Deeplabv2 \cite{deeplabv2} and Deeplabv3 \cite{deeplabv3} adopt  atrous spatial pyramid pooling to embed contextual information, which consist of parallel dilated convolutions with different dilated rates.  PSPNet \cite{PSPNet} designs a pyramid pooling module to collect the effective contextual prior, containing information of different scales.  The encoder-decoder structures \cite{refinene,ding2018context,lin2018multi,fu2017stacked} fuse mid-level and high-level semantic features to obtain different scale context.
Second, learning contextual dependencies over local features also contribute to feature representations. DAG-RNN \cite{DAGRNN}  models directed acyclic graph with recurrent neural network to capture the rich contextual dependencies. PSANet \cite{zhao2018psanet} captures  pixel-wise relation  by a convolution layer and  relative position information in spatial dimension. The concurrent work
OCNet \cite{yuan2018ocnet} adopts self-attention mechanism with ASPP to exploit context dependencies.
 In addition, EncNet \cite{encnet}  introduces a channel attention mechanism to capture global context.

\noindent{\bfseries Self-attention Modules.} Attention modules can model long-range dependencies and have been widely  applied in many tasks \cite{Atten,DiSAN,lin2016efficient,lin2017structured,tang2011image,tang2015rgb}. In particular, the work \cite{Atten} is the first to propose the self-attention mechanism to draw global dependencies of  inputs and applies it in machine translation.
Meanwhile, attention modules are increasingly applied in image vision flied. The work \cite{SelfGAN} introduces self-attention mechanism to learn a better image generator.
The work \cite{xiaolongwang2017nonlocal}, which is related to self-attention module, mainly exploring effectiveness of  non-local operation in spacetime dimension for videos and images.

Different from previous works, we extend the self-attention mechanism in the task of scene segmentation, and carefully design two types of attention modules to capture rich contextual relationships for better feature representations with intra-class compactness. Comprehensive empirical results verify the effectiveness of our proposed method.

\section{Dual Attention Network}
In this section, we first present a general framework of our network  and then introduce the two attention modules which capture long-range contextual information  in spatial and channel dimension respectively. Finally we describe how to aggregate them together for further refinement.
\subsection{Overview}
Given a picture of  scene segmentation,  stuff or objects, are diverse on scales, lighting, and views. Since convolution operations would lead to  a local receptive field, the features corresponding to  the pixels with the same label may have some differences.  These differences introduce intra-class inconsistency and affect the recognition accuracy.
To address this issue, we explore global contextual information by building associations among features with the attention mechanism. Our method  could adaptively aggregate long-range contextual information, thus improving  feature representation for scene segmentation.
 
As illustrated in Figure. \ref{DAN},  we design two types of attention modules to draw global context over local features generated by a dilated residual  network, thus obtaining  better feature representations for pixel-level prediction. We employ a pretrained residual network with the dilated  strategy \cite{deeplabv2} as the backbone. Noted that we remove the  downsampling operations and employ dilated convolutions in the last two ResNet blocks, thus enlarging the size of the final feature map size to 1/8 of the input image.  It retains more details without adding extra parameters.
Then the features from  the dilated residual network would be fed into two parallel attention modules. 
Take the spatial attention modules in the upper part of the Figure. \ref{DAN} as an example, we first apply a convolution layer to obtain the features of dimension reduction.  
Then we  feed the features into the  position attention module and generate new features of spatial long-range contextual information through the following three steps. The first step is to generate a spatial attention matrix which models the spatial relationship between any two pixels of the features. Next, we perform a matrix multiplication between the attention matrix and the original features. Third, we perform an element-wise sum operation on the above multiplied resulting matrix and original features to obtain the final representations reflecting  long-range contexts. Meanwhile, long-range contextual information in  channel dimension are captured by a channel attention module. The process of capturing the channel relationship is similar to  the position attention module except for the first step, in which channel attention matrix is calculated in  channel dimension. Finally we aggregate the outputs from the two attention modules to obtain  better feature representations for pixel-level prediction.

%-------------------------------------------------------------------------

\subsection{Position Attention Module}

Discriminant  feature representations are essential  for scene understanding, which could be obtained by capturing long-range contextual information.
However, many works \cite{PSPNet,Peng2017LargeKM} suggest that local features generated by traditional FCNs could lead to misclassification of objects and stuff.
In order to model  rich contextual relationships over local features, we introduce a position attention module.
The position attention module encodes a wider range of contextual information into local features, thus enhancing  their representation capability. 
Next, we elaborate the process to adaptively aggregate spatial contexts. 
\begin{figure}[!t]
        \centering
        \includegraphics[width=1\linewidth]{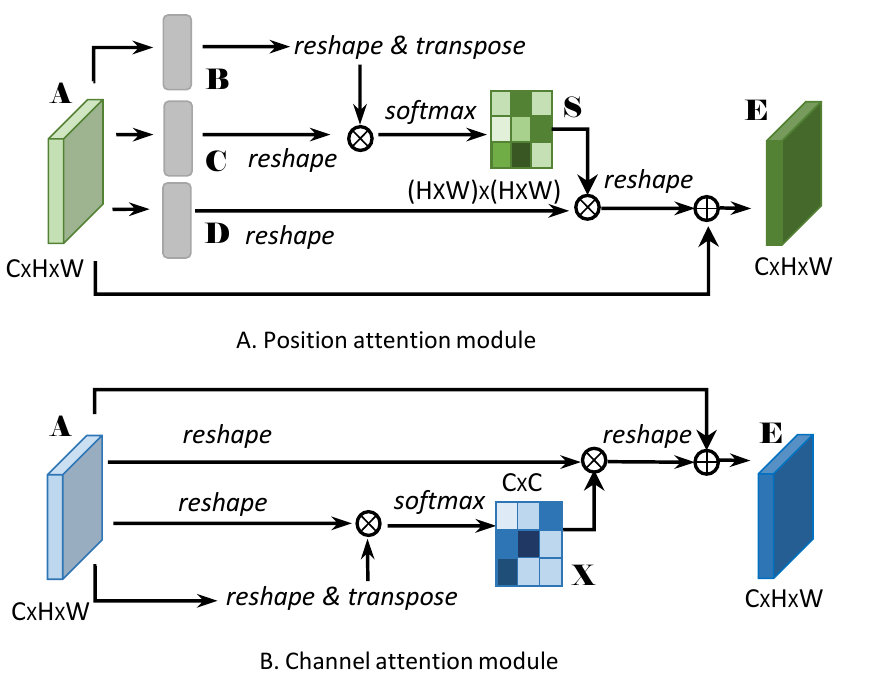}
        \caption{The details of Position Attention Module and Channel Attention Module are illustrated in (A) and (B). (Best viewed in color)}
         \label{SPAM}%
        \vspace{-1em}
\end{figure}

As illustrated in Figure.\ref{SPAM}(A), given a local feature $\mathbf{A} \in \mathbb{R}^{C \times H \times W}$,
we first feed it into a convolution layers to generate two new feature maps $\mathbf{B}$ and $\mathbf{C}$, respectively, where $\{\mathbf{B},\mathbf{C}\} \in \mathbb{R}^{C \times H \times W}$.  Then we reshape them to $ \mathbb{R}^{C \times N} $, where $N=H \times  W$ is the number of pixels.  After that we perform a matrix multiplication between the transpose of $\mathbf{C}$ and  $\mathbf{B}$, and apply a softmax layer to calculate the spatial attention map $\mathbf{S} \in \mathbb{R}^{N \times N}$:
\begin{equation}
s_{ji}=\frac{exp(B_{i}\cdot C_{j})}{\sum_{i=1}^Nexp(B_{i}\cdot C_{j})}
\end{equation}
where $s_{ji}$ measures the ${i^{th}}$ position's impact on ${j^{th}}$ position. The more similar feature representations of the two position contributes to greater correlation between them.

Meanwhile, we feed feature $\mathbf{A} $  into  a convolution layer to  generate a new feature map $\mathbf{D} \in \mathbb{R}^{C \times H \times W}$ and reshape it  to $ \mathbb{R}^{C \times N}$.
Then we  perform a matrix multiplication between $\mathbf{D}$ and  the transpose of $\mathbf{S}$ and  reshape the result to  $\mathbb{R}^{C \times H \times W}$.
Finally, we  multiply it by a scale parameter $ \alpha $ and  perform a element-wise sum operation with the features $\mathbf{A}$ to obtain the final output $\mathbf{E}\in \mathbb{R}^{C \times H \times W}$ as follows:

\begin{equation}
E_{j} =  \alpha \sum_{i=1}^N(s_{ji}D_{i} ) + A_{j}
\label{equ1}%
\end{equation}
where $\alpha$ is initialized as 0 and gradually learns to assign more weight \cite{SelfGAN}. It can be inferred from Equation \ref{equ1} that the resulting feature $\mathbf{E}$  at each position is a weighted sum of the features across all positions and original features. Therefore, it has a global contextual view and selectively aggregates contexts according to the spatial attention map. The similar semantic features  achieve mutual gains, thus imporving intra-class compact and semantic consistency.

\subsection{Channel Attention Module}
Each channel map of high level features  can be regarded as a class-specific response, and different semantic responses are associated with each other.
By exploiting the  interdependencies between channel maps, we could emphasize  interdependent feature maps and  improve the feature representation of  specific semantics.
Therefore,  we build a channel attention module to explicitly model  interdependencies between channels.

The structure of channel attention module  is illustrated in Figure.\ref{SPAM}(B). Different from the position attention module, we directly  calculate  the channel attention map $\mathbf{X} \in \mathbb{R}^{C \times C}$ from the original features  $\mathbf{A} \in \mathbb{R}^{C \times H \times W}$. Specifically,  we reshape $\mathbf{A}$ to $ \mathbb{R}^{C \times N} $, and then perform a matrix multiplication between $\mathbf{A}$ and  the transpose of $\mathbf{A}$. Finally, we apply a softmax layer to obtain the channel attention map $\mathbf{X} \in \mathbb{R}^{C \times C}$:

\begin{equation}
x_{ji}=\frac{exp(A_{i}\cdot A_{j})}{\sum_{i=1}^Cexp(A_{i}\cdot A_{j})}
\end{equation}
where $x_{ji}$ measures the ${i^{th}}$ channel's impact on the ${j^{th}}$ channel. 
In addition, we perform a matrix multiplication between  the transpose of $\mathbf{X}$ and  $\mathbf{A}$ and reshape their result to  $\mathbb{R}^{C \times H \times W}$. 
Then we multiply the result by a scale parameter $ \beta$ and  perform an element-wise sum operation with $\mathbf{A}$ to obtain the final output $\mathbf{E}\in \mathbb{R}^{C \times H \times W}$:
\begin{equation}
E_{j} =  \beta \sum_{i=1}^C(x_{ji}A_{i} ) + A_{j}
\label{equ2}
\end{equation}
where $\beta$ gradually learns a weight from 0. The Equation  \ref{equ2} shows  that the final feature of  each channel is a weighted sum of the features of  all channels and original features, which models the long-range semantic dependencies between feature maps.  It helps to boost feature discriminability. 

Noted that we do not employ convolution layers to embed features before computing relationshoips of two channels, 
since it can maintain  relationship between different channel maps. In addition, 
different from recent works \cite{encnet} which  explores channel relationships by a  global pooling or encoding layer, we  exploit spatial information at all corresponding  positions to model  channel correlations.

\subsection{Attention Module Embedding with Networks}
In order to take full advantage of long-range contextual information, we aggregate the features from these two attention modules. 
Specifically,  we transform the outputs of  two attention modules  by  a convolution layer and perform an element-wise sum to accomplish feature fusion. At last a convolution layer is followed to generate the final prediction map. We do not adopt cascading operation because it needs more GPU memory.
Noted that our attention modules  are simple and can be directly inserted in the existing FCN pipeline. They do not increase too many parameters yet strengthen feature representations effectively.
%------------------------------------------------------------------------

\section{Experiments}
To evaluate the proposed method, we carry out comprehensive experiments on Cityscapes dataset \cite{cityscapes}, PASCAL VOC2012 \cite{everingham2010pascal}, PASCAL Context dataset  \cite{pcontext} and COCO Stuff dataset \cite{cocostuff}.
Experimental results demonstrate that DANet achieves state-of-the-art performance on three  datasets. 
 In the next subsections,  we first introduce the datasets and  implementation details, then we perform a series of ablation experiments on Cityscapes dataset. Finally, we report our  results on PASCAL VOC 2012, PASCAL
Context  and COCO Stuff.

\subsection{Datasets and Implementation Details}

\noindent{\bfseries Cityscapes} The dataset  has 5,000 images captured from 50 different cities. 
Each image has ${2048\times1024}$ pixels, which have high quality pixel-level labels of 19 semantic classes.
There are 2,979 images in training set, 500 images in validation set and 1,525 images in test set. 
We do not  use coarse data in our experiments.

\begin{figure}[!t]
        \centering
        \includegraphics[width=1\linewidth]{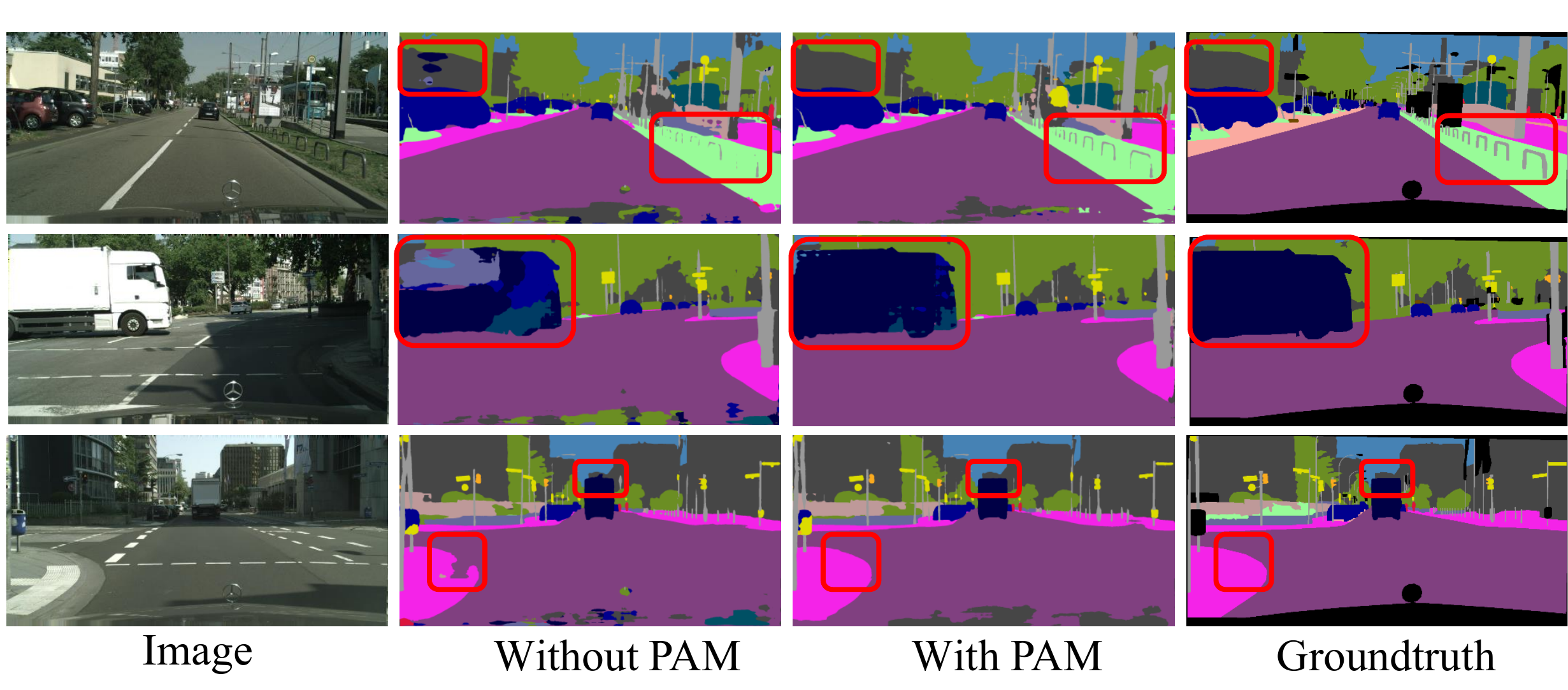}
        \vspace{-1em}
        \caption{Visualization results of position attention module on Cityscapes val set.}
         \label{PAM}%
        \vspace{0em}
\end{figure}
\begin{figure}[!t]
        \centering
        \includegraphics[width=1\linewidth]{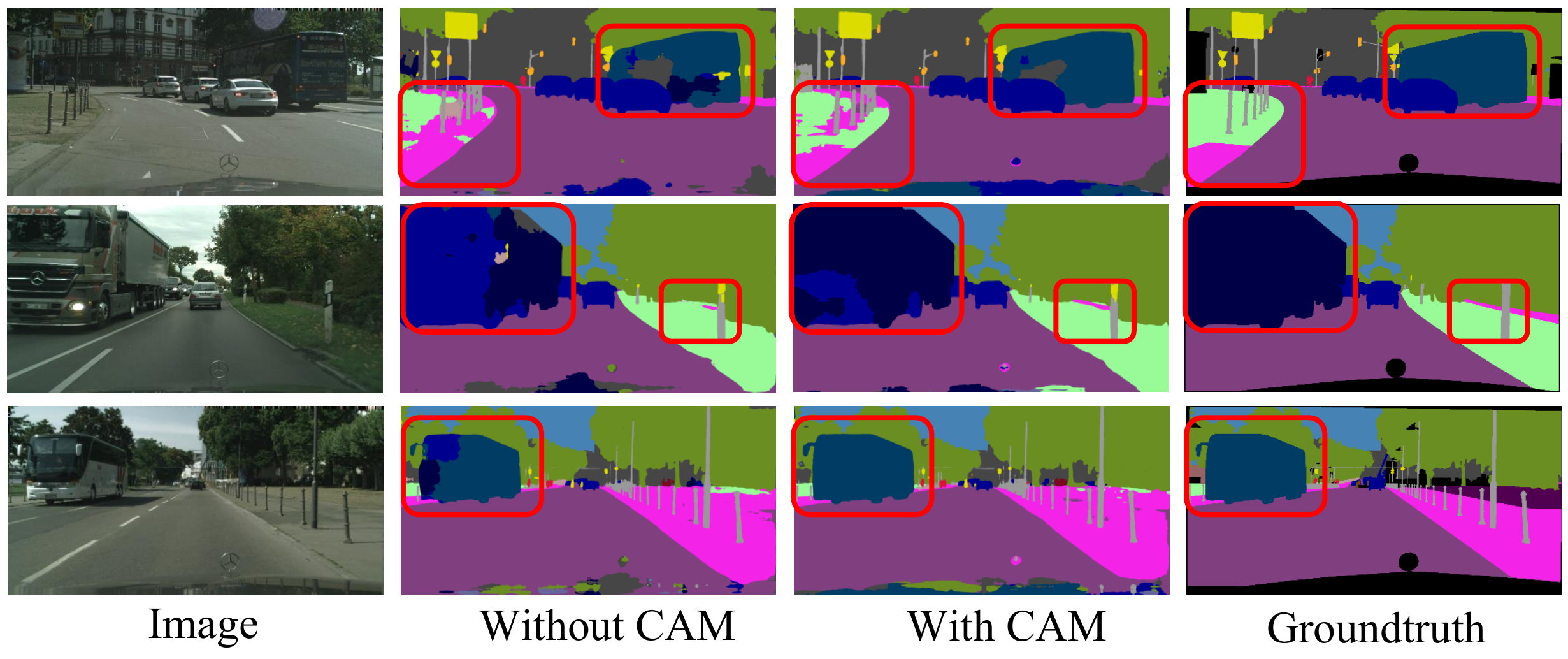}
        \vspace{-1em}
        \caption{Visualization results of channel attention module on Cityscapes val set.}
         \label{CAM}%
        \vspace{-1em}
\end{figure}

\noindent{\bfseries PASCAL VOC 2012} The dataset has 10,582 images for training, 1,449
images for validation and 1,456 images for testing, which involves 20 foreground
object classes and one background class. 

\noindent{\bfseries PASCAL Context} The dataset provides detailed semantic labels for 
whole scenes, which contains 4,998 images for training and 5,105 images for testing.
Following \cite{refinenet,encnet}, we evaluate the method on the most frequent 59
classes along with one background category (60 classes in total).

\noindent{\bfseries COCO Stuff} The dataset  contains  9,000 images  for training and 1,000 images for testing. Following \cite{refinenet,ding2018context}, we report  our results on  171 categories including 80 objects and 91 stuff annotated to each pixel. 

\subsubsection{Implementation Details}
We implement our method  based on Pytorch. Following \cite{deeplabv3,encnet}, we employ a ‘poly’ learning rate policy where
the initial learning rate is multiplied by $(1-\frac{iter}{total\_iter})^{0.9}$ after each iteration. The base learning rate is set to 0.01 for Cityscapes dataset. Momentum and weight decay  coefficients are set to 0.9 and 0.0001 respectively. We train our
model with Synchronized BN \cite{encnet}. Batchsize are set to 8 for Cityscapes and 16 for other datasets.When adopting multi-scale augmentation, we set training time to 180 epochs for COCO Stuff and 240 epochs for other datasets. Following \cite{deeplabv2}, we adopt multi-loss on the end of the network when both two attention modules are used.
For data augmentation, we apply  random cropping (cropsize 768) and random left-right flipping during
training in the ablation study for Cityscapes datasets.

%-------------------------------------------------------------------------

\subsection{Results on Cityscapes Dataset }

\subsubsection{Ablation Study for Attention Modules} We employ the dual attention modules on top of the dilation network to  capture long-range dependencies for better scene understanding. To verify  the performance of  attention modules, we conduct experiments with different settings in Table \ref{tab:ablation}.

As shown in Table \ref{tab:ablation}, the attention modules improve the performance remarkably.
Compared with the baseline FCN (ResNet-50), employing position attention module yields a result of  75.74\% in Mean IoU , which brings 5.71\% improvement.
Meanwhile, employing channel contextual module individually outperforms the baseline by 4.25\%.
When we  integrate  the two attention modules together, the performance further improves to 76.34\%.
Furthermore, when we adopt a  deeper pre-trained network (ResNet-101), the network with  two attention modules significantly improves the segmentation performance over the baseline model by 5.03\%. Results show that attention modules bring great  benefit  to scene segmentation.

The effects of  position attention modules can be visualized in Figure.\ref{PAM}. Some details and  object
boundaries are clearer with the position attention module, such as the  '‘pole’' in  the first row and the  '‘sidewalk’'  in the second row.
Selective fusion over local features enhance the discrimination of  details. 
Meanwhile, Figure.\ref{CAM} demonstrate that, with our channel attention module, some  misclassified category are now correctly classified, such as the  '‘bus’'  in the first and third row.
The selective integration among  channel maps  helps to capture context information. The semantic consistency have been improved obviously.

\begin{figure*}[!t]
        \centering
        \includegraphics[width=1.02\linewidth]{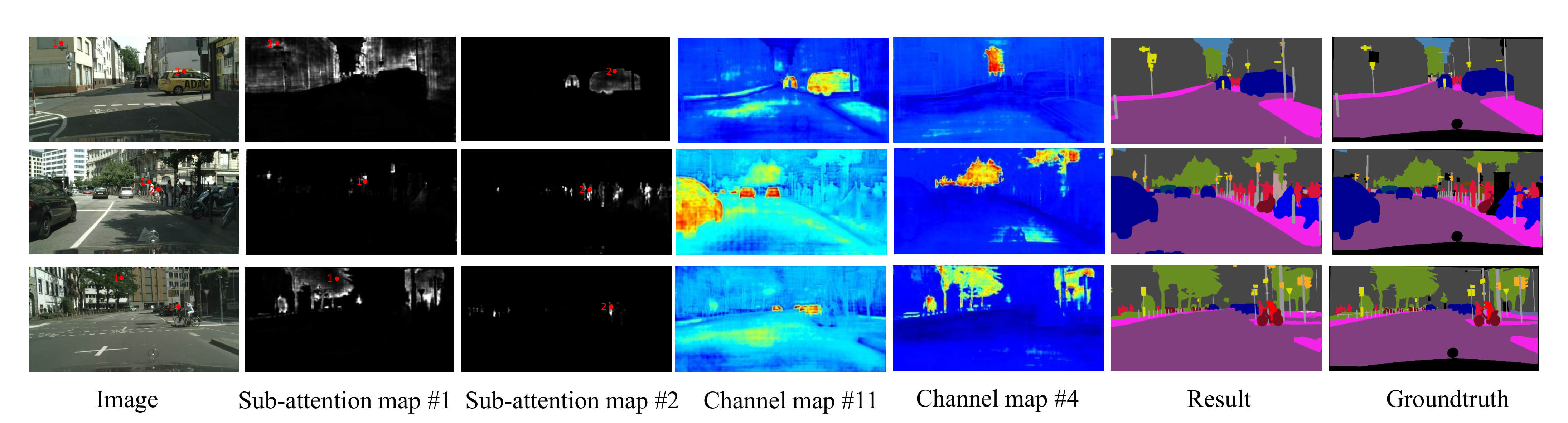}
              %  \vspace{-2em}
        \caption{Visualization results of  attention modules on Cityscapes val set. For each row, we show an input image,  two sub-attention maps $(H\times  W)$ corresponding to the ponits marked in the input image.  Meanwhile, we give two channel maps from the outputs of channel attention module, where the maps are from  $4^{th}$ and $11^{th}$ channels, respectively. Finally, corresponding result and groundtruth are provided.}
         \label{atten}%
        \vspace{-1em} 
\end{figure*}
\subsubsection{Ablation Study for Improvement Strategies}  Following \cite{deeplabv3}, we adopt the same strategies to improve performance further.
(1) DA:  Data augmentation with random scaling.
(2) Multi-Grid: we apply employ a hierarchy of grids of different sizes (4,8,16) in the last ResNet block.
(3) MS:   We average the segmentation probability maps from 8 image scales\{0.5 0.75 1 1.25 1.5 1.75 2 2.2\} for inference.

\begin{table}[t]
\centering
\resizebox{1\linewidth}{!}
  {
  \begin{tabular} {l c c c  |c c}
    \toprule[1pt]
    {\bf Method} & {\bf BaseNet} & {\bf PAM} & {\bf CAM}  &  {\bf Mean IoU\%}  \\
    \hline \hline
    Dilated FCN & Res50 & & & 70.03  \\
    DANet & Res50 & \checkmark &  & 75.74  \\
    DANet & Res50 &  &\checkmark  & 74.28  \\
    DANet & Res50 &\checkmark & \checkmark  &   76.34 \\
  \hline
    Dilated FCN & Res101 && &  72.54  \\
    DANet & Res101& \checkmark &  & 77.03  \\
    DANet & Res101&  &\checkmark  & 76.55  \\
    DANet & Res101 & \checkmark & \checkmark &  77.57  \\
    \bottomrule[1pt]
  \end{tabular}
  }
\caption{Ablation study on Cityscapes val set. {\it PAM} represents Position Attention Module, {\it CAM} represents Channel  Attention Module.  }
\label{tab:ablation}
\end{table}

\begin{table}[t]
\centering
\resizebox{\linewidth}{!}
  {
  \begin{tabular} {l c c c |c}
    \toprule[1pt]
    {\bf Method} & {\bf DA } & {\bf Multi-Grid } & {\bf MS}  &  {\bf Mean IoU\%}  \\
    \hline   \hline   
    DANet-101 &  & &&  77.57  \\
    DANet-101 & \checkmark &  & &  78.83  \\
    DANet-101 & \checkmark & \checkmark &   & 79.94  \\
    DANet-101 &\checkmark & \checkmark &\checkmark  &  81.50  \\
    \bottomrule[1pt]
  \end{tabular}
  }
\caption{Performance comparison between different strategies on Cityscape val set. {\it DANet-101} represents DANet with BaseNet ResNet-101, {\it DA} represents data augmentation  with random scaling.
 {\it Multi-Grid} represents employing multi-grid method,  {\it MS} represents multi-scale inputs during inference.}
%\vspace{-1.0em}
\label{tab:ablation2}
\end{table}

Experimental results are shown in Table \ref{tab:ablation2}. Data augmentation  with random scaling  improves the  performance by almost 1.26\%, which shows that network benefits from  enriching scale diversity of training data. 
We adopt Multi-Grid to obtain better feature representations of pretrained network, which  further achieves 1.11\% improvements.  
Finally,  segmentation map fusion further improves the performance to 81.50\%, which outperforms well-known method Deeplabv3 \cite{deeplabv3} (79.30\% on Cityscape val set) by 2.20\%.  

\begin{table*}[tp]
\renewcommand\arraystretch{1.2}
    \footnotesize
    %\scriptsize
    %\tiny
    \setlength{\tabcolsep}{4pt}
    \begin{center}
    \begin{adjustbox}{max width=\textwidth}
        \begin{tabular}{ l | c |c c c c c c c c c c c c c c c c c c c  c}
            %\hline
            \toprule[1pt]
       Methods &  \rotatebox{90}{Mean IoU} &  \rotatebox{90}{road} &  \rotatebox{90}{sidewalk} &  \rotatebox{90}{building} & \rotatebox{90}{ wall} &  \rotatebox{90}{fence} &  \rotatebox{90}{pole} & \rotatebox{90}{traffic light} &  \rotatebox{90}{traffic sign}&  \rotatebox{90}{vegetation} &  \rotatebox{90}{terrain} &  \rotatebox{90}{sky} & \rotatebox{90}{person} &  \rotatebox{90}{rider} & \rotatebox{90}{car} &  \rotatebox{90}{truck}& \rotatebox{90}{ bus}& \rotatebox{90}{ train}& \rotatebox{90}{ motorcycle}&  \rotatebox{90}{bicycle}\\
      \hline
      \hline
       DeepLab-v2\cite{deeplabv2} & 70.4 & 97.9 & 81.3 & 90.3 & 48.8 & 47.4 & 49.6 & 57.9 & 67.3 & 91.9 & 69.4 & 94.2 & 79.8 & 59.8 & 93.7 & 56.5 & 67.5 & 57.5 & 57.7 & 68.8 \\
       RefineNet~\cite{refinenet} & 73.6 & 98.2 & 83.3 & 91.3 & 47.8 & 50.4 & 56.1 & 66.9 & 71.3 & 92.3 & 70.3 & 94.8 & 80.9 & 63.3 & 94.5 & 64.6 & 76.1 & 64.3 & 62.2 & 70 \\
       GCN~\cite{Peng2017LargeKM} & 76.9 & - & - & - & - & - & - & - & - & - & - & - & - & - & - & - & - & - & - & -  \\ 
       DUC~\cite{wang2017understanding} & 77.6 & 98.5 & 85.5 & 92.8 & 58.6 & 55.5 & 65 & 73.5 & 77.9 & 93.3 & 72 & 95.2 & 84.8 & 68.5 & 95.4 & 70.9 & 78.8 & 68.7 & 65.9 & 73.8 \\  
       ResNet-38~\cite{wu2016wider} & 78.4 & 98.5 & 85.7 & 93.1 & 55.5 & 59.1 & 67.1 & 74.8 & 78.7 & 93.7 & 72.6 & 95.5 & 86.6 & 69.2 & 95.7 & 64.5 & 78.8 & 74.1 & 69 & 76.7 \\
    PSPNet~\cite{PSPNet} & 78.4 & - & - & - & - & - & - & - & - & - & - & - & - & - & - & - & - & - & - & -  \\
       BiSeNet~\cite{yu2018bisenet} & 78.9 & - & - & - & - & - & - & - & - & - & - & - & - & - & - & - & - & - & - & -  \\ 
       PSANet~\cite{zhao2018psanet} & 80.1 & - & - & - & - & - & - & - & - & - & - & - & - & - & - & - & - & - & - & -  \\ 
       DenseASPP~\cite{yang2018denseaspp} & 80.6 & \textbf{98.7} & \textbf{87.1} & 93.4 & \textbf{60.7} & 62.7 & 65.6 & 74.6 & 78.5 & 93.6 & 72.5 & 95.4 & 86.2 & 71.9 & 96.0 & \textbf{78.0} & \textbf{90.3} & 80.7 & 69.7 & 76.8\\         
 \hline 
       DANet & \textbf{81.5} & 98.6 & 86.1 & \textbf{93.5} & 56.1 & \textbf{63.3} & \textbf{69.7} & \textbf{77.3} & \textbf{81.3} & \textbf{93.9} & \textbf{72.9} & \textbf{95.7} & \textbf{87.3} & \textbf{72.9} & \textbf{96.2} & 76.8 & 89.4 & \textbf{86.5} & \textbf{72.2} & \textbf{78.2}\\     
        \bottomrule[1pt]
        \end{tabular}
    \end{adjustbox}
    \end{center}
    \vspace{-1.5em}
    \caption{Per-class results on Cityscapes testing set. DANet outperforms existing approaches and achieves 81.5\% in Mean IoU. }
\vspace{-1.5em}
\label{tab:cityset}
\end{table*}
\vspace{-0.8em}
%-------------------------------------------------------------------------
\
\subsubsection{Visualization of Attention Module}
For position attention, the overall self-attention map is in size of $(H\times  W) \times (H \times W)$, which means that for each specific point in the image, there is an corresponding sub-attention map whose size is $(H\times  W)$. In Figure.\ref{atten}, for each input image, we select two points (marked as ${\#1}$ and ${\#2}$) and show their corresponding sub-attention map in columns 2 and 3 respectively.
We observe that the position attention module could capture clear semantic similarity and long-range relationships. 
For example, in  the first row, 
the red point ${\#1}$ are marked on a building and its attention map (in column 2) highlights most the areas where the buildings lies on. Moreover, in the sub-attention map, the boundaries are very clear even though  some of them are far away  from the point ${\#1}$. As for the point ${\#2}$, its attention map focuses on most positions labeled as "car". 
In the second row, the same holds for the '‘traffic sign’' and   '‘person’' in global region, even though the number of corresponding pixels is less.  
The third row is for the  '‘vegetation’'  class and '‘person’' class. In particular, the point ${\#2}$ does not respond to the nearby  '‘rider'‘ class, but it does respond to the '‘person’'  faraway.

For channel attention, it is hard to give comprehensible visualization about the attention map directly. Instead, we show some attended channels to see whether they highlight clear semantic areas. In Figure.\ref{atten}, we display the eleventh and fourth attended channels in column 4 and 5. We find that the  response of specific semantic is noticeable after  channel attention module enhances.
For example,  $11^{th}$ channel map  responds to the '‘car'‘  class in all three examples, and  $4^{th}$ channel map is for  the '‘vegetation'‘ class, which benefits for the segmentation of  two scene categories.
In short, these visualizations  further demonstrate the necessity of capturing long-range dependencies for improving feature representation in scene segmentation.

\subsubsection{Comparing with State-of-the-art} 
We further compare our method with existing methods on the
Cityscapes testing set. Specifically,  we train our DANet-101 with only fine annotated data and submit our test results to the official
evaluation server. Results are shown in Table \ref{tab:cityset}.
DANet outperforms existing approaches with dominantly advantage.
In particular, our model  outperforms PSANet\cite{zhao2018psanet} by a large margin with the same backbone ResNet-101. Moreover, it also surpasses DenseASPP\cite{yang2018denseaspp}, which use  more powerful pretrained models than ours.

\begin{table}[t]
\centering
\resizebox{1\linewidth}{!}
  {
  \begin{tabular} {l c c c  |c c}
    \toprule[1pt]
    {\bf Method} & {\bf BaseNet} & {\bf PAM} & {\bf CAM}  &  {\bf Mean IoU\%}  \\
    \hline \hline
    Dilated FCN & Res50 & & & 75.7  \\
    DANet & Res50 & \checkmark & \checkmark  & 79.0  \\
    DANet & Res101 &\checkmark & \checkmark  & 80.4 \\
    \bottomrule[1pt]
  \end{tabular}
  }
\vspace{0.2em}
\caption{Ablation study on PASCAL  VOC 2012 val set. {\it PAM} represents Position Attention Module, {\it CAM} represents Channel  Attention Module.  }
\vspace{-1em}
\label{tab:pvocval}
\end{table}

\begin{table}[t]
\centering
\resizebox{1\linewidth}{!}
  {
  \begin{tabular} {l |c  }
    \toprule[1pt]
    {\bf Method} & {\bf Mean IoU\%}  \\
    \hline \hline
    FCN \cite{FCN} &  62.2 \\
    DeepLab-v2(Res101-COCO) \cite{deeplabv2} & 71.6 \\ 
    Piecewise\cite{lin2016efficient} & 75.3 \\ 
    ResNet38 \cite{refinenet} &  82.5\\
    PSPNet(Res101)  \cite{PSPNet} &  82.6\\ 
    EncNet (Res101) \cite{encnet} & \textbf{ 82.9} \\
	\hline	
       \hline
        DANet(Res101) & 82.6\\
    \bottomrule[1pt]
  \end{tabular}
  }
\vspace{0.2em}
\caption{Segmentation results on PASCAL VOC 2012  testing set.  }
\vspace{-0.8em}
\label{tab:poctest}
\end{table}

%------------------------------------------------------------------------

\subsection{Results on PASCAL VOC 2012 Dataset }
We carry out experiments on the PASCAL VOC 2012 dataset to further evaluate the effectiveness of our method.  
Quantitative results of PASCAL VOC 2012 val set are shown in Table. \ref{tab:pvocval}. Our attention modules improves performance significantly, where DANet-50  exceeds the baseline by 3.3\%.  When we adopt a deeper  network ResNet-101, the model further achieves a Mean IoU of  80.4\%.  Following \cite{PSPNet,deeplabv3, encnet}, we employ the PASCAL VOC 2012 trainval set further fine-tune our best model. The results of PASCAL VOC2012 on test set is  are shown in Table \ref{tab:poctest}. 

\vspace{0.2cm}
\begin{table}[t]
\centering
\resizebox{1\linewidth}{!}
  {
  \begin{tabular} {l |c  }
    \toprule[1pt]
    {\bf Method} & {\bf Mean IoU\%}  \\
    \hline \hline

    FCN-8s \cite{FCN} &  37.8 \\
    Piecewise \cite{lin2016efficient}  & 43.3\\ 
    DeepLab-v2 (Res101-COCO) \cite{deeplabv2} & 45.7 \\ 
    RefineNet (Res152) \cite{refinenet} &  47.3\\ 
    PSPNet (Res101) \cite{PSPNet} &  47.8\\
    Ding et al.( Res101) \cite{ding2018context} &  51.6 \\
   EncNet  (Res101) \cite{encnet} &  51.7 \\
	\hline	
       \hline
        Dilated FCN(Res50)   &  44.3 \\
        DANet (Res50) &  50.1 \\
        DANet (Res101) & \textbf{52.6} \\
    \bottomrule[1pt]
 \end{tabular}
  }
\caption{Segmentation results on PASCAL Context  testing set.}
\vspace{-0.8em}
\label{tab:pcontext}
\end{table}
\vspace{-0.8em}

\subsection{Results on PASCAL Context Dataset }
In this subsection, we carry out experiments on the PASCAL Context dataset to further evaluate the effectiveness of our method.
We adopt the same training  and testing settings on  PASCAL VOC 2012 dataset.
Quantitative results of PASCAL Context are shown in Table. \ref{tab:pcontext}.
The baseline (Dilated FCN-50) yields  Mean IoU 44.3\%. DANet-50 boosts the performance
to 50.1\%. Furthermore, with a deep pretrained network ResNet101, our model results  achieve Mean IoU 52.6\%, which outperforms previous methods by a large margin.
Among previous works, Deeplab-v2 and RefineNet adopt  multi-scale feature fusion by different atrous convolution or different stage of encoder.
In addition, they trained their model with extra COCO data or adopt a deeper model (ResNet152) to improve their segmentation results. 
Different from the previous methods, we introduce attention modules to capture global dependencies explicitly, and the proposed method can achieve better performance. %Some examples on  PASCAL Context  val set are shown in Figure \ref{}

\subsection{Results on COCO Stuff  Dataset }

\begin{table}[t]
\centering
%\resizebox{1\linewidth}{!}
  {
  \begin{tabular} {l| c }
    \toprule[1pt]
    {\bf Method} & {\bf Mean IoU\%} \\
    \hline \hline
    FCN-8s \cite{FCN} &  22.7 \\
DeepLab-v2(Res101) \cite{deeplabv2} &  26.9 \\ 
DAG-RNN \cite{DAGRNN}  &  31.2\\
RefineNet (Res101) \cite{refinenet}  &  33.6\\
Ding et al.( Res101) \cite{ding2018context}&35.7 \\
	\hline	
       \hline
        Dilated FCN (Res50) &31.9 \\
        DANet (Res50) & 37.2\\
        DANet (Res101) & \textbf{39.7} \\
    \bottomrule[1pt]
  \end{tabular}
  }
\caption{Segmentation results on COCO Stuff testing set.}
\vspace{-0.8em}
%\vspace{-1em}
\label{tab:coco}
\end{table}

We also conduct experiments on the COCO Stuff dataset to verify the generalization of our proposed network.  
Comparisons with previous state-of-the-art methods are reported in Table. \ref{tab:coco}. Results show that our model achieves 39.7\% in Mean IoU, which outperforms these methods  by a large margin. Among the compared methods, DAG-RNN \cite{DAGRNN} utilizes  chain-RNNs for 2D images to model rich spatial dependencies, and Ding et al. \cite{ding2018context}  adopts a gating mechanism in the decoder stage for improving inconspicuous objects and background stuff segmentation.  our method could  capture  long-range contextual information more effectively and learn better feature representation in scene segmentation.

\section{Conclusion}

In this paper, we have presented a Dual Attention Network (DANet) for scene
segmentation, which adaptively integrates local semantic features using the self-attention mechanism. 
Specifically, we introduce a position attention module and a channel attention module to  
capture global dependencies in the spatial and channel dimensions respectively. 
The ablation experiments show that dual attention modules capture long-range contextual information effectively and give more precise segmentation results.
Our  attention network achieves outstanding performance consistently on four scene segmentation datasets, i.e. Cityscapes, Pascal VOC 2012, Pascal Context, and COCO Stuff. In addition, it is important to decrease the computational complexity and enhance the robustness of the model, which will be studied in future work.

\section*{Acknowledgment}
This work was supported by Beijing Natural Science Foundation (4192059) and National Natural Science Foundation of China (61872366, 61472422 and 61872364). 

{\small
\bibliographystyle{ieee_fullname}
\bibliography{references}
}

\end{document}